\documentclass[lettersize,journal]{IEEEtran}
\usepackage{amsmath,amsfonts}
\usepackage{algorithmic}
\usepackage{algorithm}
\usepackage{array}
\usepackage[caption=false,font=normalsize,labelfont=sf,textfont=sf]{subfig}
\usepackage{textcomp}
\usepackage{stfloats}
\usepackage{url}
\usepackage{verbatim}
\usepackage{graphicx}
\usepackage{cite}
\begin{document}

\title{QuatE-D: A Distance-Based Quaternion Model for Knowledge Graph Embedding}

\author{
 Hamideh-Sadat~Fazael-Ardakani$^{\ddag}$, Hamid~Soltanian-Zadeh$^{\ddag}$
 \\
\small $^\ddag$School of Electrical and Computer Engineering., College of Engineering., University of Tehran,Tehran., Iran. \\

}



\maketitle

\begin{abstract}
Knowledge graph embedding (KGE) methods aim to represent entities and relations in a continuous space while preserving their structural and semantic properties. Quaternion-based KGEs have demonstrated strong potential in capturing complex relational patterns. In this work, we propose QuatE-D, a novel quaternion-based model that employs a distance-based scoring function instead of traditional inner-product approaches. By leveraging Euclidean distance, QuatE-D enhances interpretability and provides a more flexible representation of relational structures. Experimental results demonstrate that QuatE-D achieves competitive performance while maintaining an efficient parameterization, particularly excelling in Mean Rank reduction. These findings highlight the effectiveness of distance-based scoring in quaternion embeddings, offering a promising direction for knowledge graph completion.
\end{abstract}


\section{Introduction}
\IEEEPARstart{K}{nowledge Graphs} (KGs) are structured representations of real-world knowledge, expressed as triples (head, relation, tail) that denote relationships between entities. These graphs encapsulate factual information about entities, such as objects, events, or abstract concepts, and their interconnections. KGs have emerged as foundational tools in a wide range of applications, including question-answering\cite{hao2017end,cui2019kbqa,saxena2020improving}, natural language processing\cite{yang2019leveraging}, and recommendation systems\cite{zhang2016collaborative,wang2018dkn}. Their ability to represent and infer complex relationships makes them indispensable for semantic reasoning and downstream AI applications. However, real-world KGs are often incomplete, which limits their effectiveness in various tasks.

To address the incompleteness of KGs, Knowledge Graph Embedding (KGE) methods have gained significant attention. KGE encodes entities and relations into low-dimensional vector spaces, capturing intricate relational patterns and enabling efficient reasoning. By facilitating the prediction of missing links in KGs, KGE enhances their utility and completeness. These embeddings have proven effective in improving performance across various knowledge-intensive applications, making them a critical tool in modern AI research and development.

Existing KGE methods explore a variety of embedding spaces and scoring functions to capture the diverse relational patterns in KGs. For instance, translational models like TransE\cite{bordes2013translating} represent relations as translations in real-valued space but struggle with symmetry patterns. Rotational models, such as RotatE\cite{sun2019rotate}, leverage complex space rotations to model symmetry, antisymmetry, inversion, and composition. Extending further, QuatE\cite{zhang2019quaternion} employs the quaternion space with Hamilton products to encode richer interactions between entities and relations. Despite their successes, these models often rely on inner-product-based scoring functions, which may limit their ability to generalize across all relation types and capture nuanced geometric properties.

In this paper, we introduce QuatE-D (Quaternion embedding with Distance Scoring), a novel approach to KGE that leverages the expressive power of quaternion space while departing from traditional inner-product-based scoring functions. Instead, QuatE-D employs an Euclidean distance-based scoring function, offering a fresh perspective on modeling relational patterns.

Our model addresses several key limitations in existing quaternion-based embeddings. While previous methods like QuatE use inner-product scoring, which can struggle with certain relational patterns, QuatE-D's distance-based approach provides a more geometrically interpretable and flexible framework. Furthermore, QuatE-D demonstrates strong generalization capabilities and multi-level improvements in embedding dimensions.For instance, when tested on WN18 with an embedding dimension of 50 (one-sixth of QuatE's dimension), our model achieves better Mean Rank (MR) and Hit@10 scores than QuatE.
Extensive experiments on benchmarks like FB15K\cite{bollacker2008freebase}, and WN18\cite{miller1995wordnet} further validate our method. Results show that QuatE-D outperforms state-of-the-art approaches across diverse datasets, proving its efficacy in accurately modeling relational patterns and predicting missing links in knowledge graphs.
\section{Related work}

Knowledge graph embedding (KGE) models are designed to represent entities and relationships within low-dimensional vector spaces. This transformation facilitates efficient reasoning for tasks like link prediction, entity classification, and recommendations. Throughout the years, various KGE methods have emerged, each differing in their embedding spaces, scoring functions, and their capability to model relational patterns. Generally, these methods fall into three categories: Translational Models, Semantic-Based Models, and Neural Network-Based Models.

\subsection{Translational Models}
Translational models interpret relations as translations in the embedding space, aiming to minimize a distance-based objective for triples $(h,r,t)$. The pioneering TransE\cite{bordes2013translating} uses the scoring function:
\begin{align}
f(h,r,t) = -\|\mathbf{h}+\mathbf{r}-\mathbf{t}\|_2
\end{align}
where  $ \mathbf{h}$, $\mathbf{r}$, and $\mathbf{t} $ are the embeddings of the head entity, relation, and tail entity, respectively, despite its simplicity and efficiency, TransE struggles with one-to-many and symmetric relations. Subsequent models have extended TransE to address its limitations. TransH\cite{wang2014knowledge} introduced relation-specific hyperplanes to improve representation flexibility. In contrast, TransR\cite{lin2015learning} embeds entities and relations in distinct spaces, enabling richer relational modeling. To account for relation and entity heterogeneity, TransD\cite{ji2015knowledge} dynamically learns mapping matrices.
 
 RotatE\cite{sun2019rotate} represents relations as rotations in the complex plane with a scoring function:
\begin{align}
f(h,r,t) = -\|\mathbf{h}\circ\mathbf{r}-\mathbf{t}\|_2
\end{align}
where $\circ$ denotes the Hadamard product. RotatE effectively models both symmetric and antisymmetric relations. Similarly, Rot-Pro\cite{chao2020pairre} enhances relation modeling by combining rotations and projections. These models leverage rotational transformations to capture relational intricacies better.
Some models extend beyond the traditional Euclidean space to address different relational patterns. For example, MuRP\cite{balazevic2019multi} maps embeddings into hyperbolic space, which preserves hierarchical structures in knowledge graphs. Additionally, TorusE\cite{ebisu2019generalized} models embeddings on a torus to mitigate boundary effects commonly found in translational approaches. Furthermore, PairRE\cite{chao2020pairre} uses independent translation vectors for head and tail entities, offering greater flexibility in modeling relational constraints.
\subsection{Semantic-Based Models}
Semantic-based models utilize bilinear or inner-product scoring functions to capture interactions between entities and relations. RESCAL\cite{nickel2011three}
is one of the earlier models in this category, which factorizes the adjacency tensor of the knowledge graph using bilinear mappings. The scoring function is:
\begin{align}
f(h,r,t) = \mathbf{h}^T\mathbf{R}_r\mathbf{t},
\end{align}
where $\mathbf{R}_r $ is a relation-specific matrix. Due to the dense matrix $\mathbf{R}_r$ RESCAL can capture complex interactions but may suffer from scalability issues. To address these concerns, DistMult \cite{yang2014embedding}‌simplifies the bilinear product by using a diagonal relation matrix, represented by:
\begin{align}
f(h,r,t) = \mathbf{h}^T{\rm diag}(\mathbf{r})\mathbf{t}
\end{align}
This reduces computational complexity but assumes symmetric relations. ANALOGY \cite{liu2017analogical} extends these ideas by leveraging analogical structures between entities, enhancing interpretability without a unique mathematical formula distinct from the models mentioned above.

ComplEx\cite{trouillon2016complex} addresses the limitation of symmetric assumptions by extending the model into the complex space, handling asymmetric relations effectively with the scoring function:
\begin{align}
f(h,r,t) = Re(<\mathbf{r},\mathbf{h},\overline{\mathbf{t}}>) 
\end{align}
where $\overline{\mathbf{t}}$ is the complex conjugate of $\mathbf{t}$. HolE\cite{nickel2016holographic} enhances efficiency and expressiveness by combining circular correlation with tensor factorization.QuatE\cite{zhang2019quaternion}
adopts quaternion embeddings and models interactions using the Hamilton product, represented by:
\begin{align}
f(h,r,t)=\mathbf{Q}_h \otimes \mathbf{W}^{\triangleleft}_r \cdot \mathbf{Q}_t
\end{align}
where $\mathbf{Q}_h$, $\mathbf{W}_r$, and $\mathbf{Q}_t$ are the quaternion embedding of head, relation, and tail. This approach captures multi-relational patterns and enhances geometric alignment. DualE\cite{cao2021dual} and BiQUE\cite{guo2021bique}  models further extend this by using dual and biquaternion algebra to capture complex relational structures.

Recent advancements in quaternion-based KGEs integrate additional mechanisms to enhance representation learning. \cite{liang2024effective} incorporates convolutional layers for improved feature extraction, while QuatDE \cite{gao2021quatde} introduces dynamic quaternion embeddings for adaptability over time. Hierarchy-Aware Quaternion Embeddings \cite{liang2024hierarchy} enforce hierarchical constraints to improve reasoning, and QuatSE \cite{li2022quatse} utilizes spherical linear interpolation for optimized transformations. Quatre enhances relation-aware modeling through quaternion transformations. These advancements showcase the growing importance of quaternion-based representations in capturing complex relational patterns, leading to improved performance in knowledge graph completion and reasoning tasks.
\subsection{Neural Network-Based Models}
Neural network-based approaches offer flexibility in modeling complex, non-linear patterns within KGs. SME\cite{bordes2014semantic} combines structured semantic embeddings to represent entities and relations effectively. 
Multi-layer perceptrons (MLPs) are used in models like MLP\cite{dong2014knowledge}
to encode entity and relation features. ConvE\cite{dettmers2018convolutional} applies 2D convolutions to embeddings for expressive feature learning, whereas ConvKB \cite{nguyen2017novel} uses 1D convolutions over concatenated embeddings to capture interaction patterns. Graph convolutional networks (GCNs) play a crucial role in models like R-GCN\cite{schlichtkrull2018modeling}, which aggregate neighborhood information to improve relational reasoning. CompGCN\cite{vashishth2019composition} further enhances this by combining relational and entity embeddings during graph convolution for enriched representations.
Moreover, KG-BERT\cite{yao2019kg} leverages pre-trained transformers to contextualize and reason over KG triples, providing a powerful mechanism for understanding complex relational patterns.
In this work, we propose QuatE-D (Quaternion Embedding with Distance Scoring), an advancement over existing quaternion-based models. Unlike prior work such as QuatE, which relies on an inner-product scoring function, QuatE-D employs an Euclidean distance-based scoring function.
\begin{table}[t]
    \centering
    \caption{Representative Models and Scoring Functions}
    \resizebox{0.5\textwidth}{!}{\begin{tabular}{|c|c|c|c|}
         \hline Model & Category& Embedding Space & Scoring Function   \\
         \hline
         TransE\cite{bordes2013translating} & Translation & $ \mathbf{h},\mathbf{r}, \mathbf{t} \in \mathbb{R}^d$ & $-\mid \mathbf{h}+\mathbf{r}-\mathbf{t} \|_{1 / 2}$ \\
         \hline 
         {RotatE}\cite{sun2019rotate} & Translation&  $\mathbf{h}, \mathbf{r}, \mathbf{t} \in \mathbb{C}^d$ & $-\|\mathbf{h}\circ\mathbf{r}-\mathbf{t}\|$ \\
         \hline 
         DistMult\cite{yang2014embedding} & Semantic-Based & $\mathbf{h}, \mathbf{r}, \mathbf{t} \in \mathbb{R}^d$ &  $\mathbf{h}^{T} \operatorname{diag}(\mathbf{r}) \mathbf{t}$ \\
         \hline 
         ComplEx\cite{trouillon2016complex} & Semantic-Based & $\mathbf{h}, , \mathbf{r}, \mathbf{t} \in \mathbb{C}^d$ & $\operatorname{Re}(\langle\mathbf{r}, \mathbf{h}, \overline{\mathbf{t}}\rangle)$ \\
         \hline 
         QuatE\cite{zhang2019quaternion} & Semantic-Based & $Q_h,W_r, Q_t \in \mathbb{H}^d$ & $Q_h \otimes W_r^{\triangleleft} \cdot Q_t$  \\
         \hline 
    \end{tabular}}
    \label{tab:my_label}
\end{table}
\section{Background}
Quaternions, introduced by William Rowan Hamilton in 1843, extend complex numbers to four-dimensional space. Represented as a combination of one real and three imaginary components \cite{hamilton1844ii}. A quaternion $q$ is expressed as:
\begin{align}
    q =  q_0+q_1i+q_2j+q_3k
\end{align}
where $q_0,q_1,q_2,q_3 \in \mathbb{R}$ are real numbers, and $i,j,k$ are imaginary units that satisfy the following multiplication rules:
\begin{align}
& i^2=j^2=k^2=ijk=-1 \nonumber \\
&ij=-ji=k\nonumber \\
&jk=-jk=i\nonumber \\
&ki=-ki=j
\end{align}
The quaternion $q$ can also be expressed in the form:
\begin{align}
    q = (q_0 , \mathbf{w})
\end{align}
where $q_0$ is the scaler part and $\mathbf{w}=(q_1,q_2,q_3)$ is the vector part. This representation is particularly useful for understanding quaternion's geometric and algebraic properties.

Quaternions offer a rich algebraic framework with operations such as addition, multiplication, conjugation, and norm calculation, making them suitable for representing rotations, relational structures, and embeddings. For example, given two quaternions:
\begin{align}
    \mathbf{q}= (q_0 , \mathbf{w})~~~,~~~\mathbf{p}= (p_0 , \mathbf{v}) 
\end{align}
\begin{itemize}
    \item {Quaternion Addition and Subtraction:}
    \begin{align}\label{add}
\mathbf{p} \pm \mathbf{q} = (p_0\pm q_0 , \mathbf{v}\pm \mathbf{w})
\end{align}
Quaternion addition and subtraction are performed component-wise.
\item {Conjugation:}
The conjugate of a quaternion $q$ is 
\begin{align}
    q^* = (q_0 , -\mathbf{w}) = q_0-q_1i-q_2j-q_3k.
\end{align}
\item {Norm:}
The norm of a quaternion $q=q_0+q_1i+q_2j+q_3k$ is given by: 
\begin{align}\label{norm}
\|q\| = qq^*= q_0^2+q_1^2+q_2^2+q_3^2
\end{align}
\item {Dot Product:}
The dot product between two quaternions is defined as:
\begin{align}\label{dotproduct}
p\cdot q = p_0q_0+p_1q_1+p_2q_2+p_3q_3
\end{align}
\item{Multiplication(Hamilton Product):}
The Hamilton product  of two quaternions is defined as:
\begin{align}\label{product}
& p \otimes q =  (p_0q_0-\mathbf{v}.\mathbf{w} , p_0\mathbf{w}+q_0\mathbf{v}+\mathbf{v}\times\mathbf{w}) \nonumber \\
&=(p_0q_0- p_1q_1-p_2q_2-p_3q_3) \nonumber \\
&+(p_0q_1+p_1q_0+p_2q_3-p_3q_2)i\nonumber \\
&+ (p_0q_2+p_2q_0+p_3q_1-p_1q_3)j \nonumber \\
&+(p_0q_3+p_3q_0+p_1q_2-p_2q_1)k 	
\end{align}
where $\mathbf{v}.\mathbf{w}$ and $\mathbf{v}\times\mathbf{w}$ are inner and cross product of vectors,respectively\cite{kenwright2023survey}.
\end{itemize}

\section{Problem statement}


Knowledge Graph Embeddings (KGE) are powerful tools for modeling and inferring complex relationships within knowledge graphs, essential for tasks such as link prediction, entity resolution, and semantic search. However, traditional KGE methods often struggle with accurately representing diverse relational patterns like symmetry, inversion, and composition. Quaternion-based embeddings offer a promising solution due to their capacity to encode richer relational information in lower-dimensional spaces.

In this work, we introduce a novel Quaternion Knowledge Graph Embedding model, QuatE-D, which employs a Euclidean distance-based scoring function inspired by the RotatE framework. Unlike conventional models that rely on dot-products, QuatE-D utilizes the Euclidean distance metric to effectively capture relational patterns. This new approach addresses the limitations of existing methods, ensuring accurate modeling of critical properties such as symmetry, inversion, and composition. We detail the framework's design and notation, followed by an analysis of its advantages.

Let $ G$ represent a knowledge graph containing  $N $ entities and 
$M$ relations. Entity embeddings are modeled as a quaternion matrix $\mathbf{Q}\in \mathbb{H}^{N \times k}$, where each row corresponds to a quaternion embedding of dimensionality $k$. Relation embeddings are represented similarly as a quaternion matrix $\mathbf{W} \in \mathbb{H}^{M \times k}$.

For a triple $(h,r,t)$, the quaternion embedding of the head and tail entity are denoted by 
\begin{align}
&\mathbf{Q}_h =\{\mathbf{a}_h+\mathbf{b}_hi+\mathbf{c}_hj+\mathbf{d}_hk:\mathbf{a}_h,\mathbf{b}_h,\mathbf{c}_h,\mathbf{d}_h \in \mathbb{R}^k\} \nonumber \\
& \mathbf{Q}_t = \{\mathbf{a}_t+\mathbf{b}_ti+\mathbf{c}_tj+\mathbf{d}_tk : \mathbf{a}_t,\mathbf{b}_t,\mathbf{c}_t,\mathbf{d}_t \in \mathbb{R}^k\} 
\end{align}
The relation $r$ is embedded as
\begin{align}
      \mathbf{W}_r=\{\mathbf{a}_r+\mathbf{b}_ri+\mathbf{c}_rj+\mathbf{d}_rk:\mathbf{a}_r,\mathbf{b}_r,\mathbf{c}_r,\mathbf{d}_r \in \mathbb{R}^k\} 
\end{align}

 To ensure robustness and eliminate the effect of scaling, we normalize the relation quaternion
\begin{align}
  \mathbf{W}_r^{\triangleleft}(\mathbf{p},\mathbf{q},\mathbf{u},\mathbf{v}) = \frac{\mathbf{W}_r}{|\mathbf{W}_r|} = \frac{\mathbf{a}_r+\mathbf{b}_ri+\mathbf{c}_rj+\mathbf{d}_rk}{\sqrt{\mathbf{a}_r^2+\mathbf{b}_r^2+\mathbf{c}_r^2+\mathbf{d}_r^2}}
\end{align}
This step guarantees that $\|\mathbf{W}_r^{\triangleleft}  \| = 1$ reduces the degrees of freedom and ensures stable embeddings.
To model relations, the head entity  $\mathbf{Q}_h $ undergoes a rotation transformation. This transformation uses the Hamilton product to rotate $\mathbf{Q}_h$ by the normalized relation quaternion $\mathbf{W}_r^{\triangleleft}$ as follows:
\begin{align}\label{hamiltonKGE}
    &\mathbf{Q}'_h(\mathbf{a}'_h,\mathbf{b}'_h,\mathbf{c}'_h,\mathbf{d}'_h) = \mathbf{Q}_h \otimes  \mathbf{W}_r^{\triangleleft}=\nonumber \\
    & (\mathbf{a}_h \circ \mathbf{p} - \mathbf{b}_h \circ \mathbf{q} - \mathbf{c}_h \circ \mathbf{u} - \mathbf{d}_h \circ \mathbf{v}) \nonumber \\
    &+(\mathbf{a}_h \circ \mathbf{q} +\mathbf{b}_h \circ \mathbf{p} + \mathbf{c}_h \circ \mathbf{v}- \mathbf{d}_h \circ \mathbf{u})i\nonumber \\
    &+ (\mathbf{a}_h \circ \mathbf{u}- \mathbf{b}_h \circ \mathbf{v} + \mathbf{c}_h \circ \mathbf{p} + \mathbf{d}_h \circ \mathbf{q})j\nonumber\\
    &+(\mathbf{a}_h \circ \mathbf{v}+ \mathbf{b}_h \circ \mathbf{u} - \mathbf{c}_h \circ \mathbf{q} + \mathbf{d}_h \circ \mathbf{p})k
    \end{align}
where $\circ$ denotes the element-wise multiplication between two vectors.

The transformed head entity quaternion $\mathbf{Q}'_h$ is then compared to $\mathbf{Q}_t$ using the quaternion Euclidean distance:
\begin{align}\label{score}
&\phi(h,r,t)=\|\mathbf{Q}'_h-\mathbf{Q}_t\|_2= \nonumber\\
&=\sqrt{(a'_h-a_t)^2+(b'_h-b_t)^2+(c'_h-c_t)^2+(d'_h-d_t)^2}
\end{align}
To optimize the distance-based scoring model, we employ the Margin Ranking Loss, which ensures positive triplets have lower scores than negative triplets by a margin $\gamma$. Let $Y_{hrt} \in \{1,-1\}$  denote the label for a triplets, where $Y_{hrt} = 1 $ represents a positive triplet $ (h,r,t) $ and $Y_{hrt} =-1$ represents a negative triplet $(h',r,t')$.
The loss function is formulated as:
\begin{align}
   L(Q,W) = \max(0,\gamma+Y_{hrt}\phi(h,r,t))
\end{align}
For a positive triplet ($Y_{hrt} = 1$) , the loss ensures that $\phi(h,r,t)+\gamma \le \phi(h',r,t')$, minimizing the distance between entities connected by observed relationships. Conversely, for a negative triplet ($Y_{hrt} = -1$) , the loss enforces $\phi(h,r,t) \ge \gamma +\phi(h',r,t')$, increasing the separation for unobserved triplets.

To mitigate overfitting, we add regularization terms on the entity embeddings  and relation embeddings. The complete loss function becomes:
\begin{align}
&L(Q,W) = \sum_{(h,r,t) \cup (h',r,t')} \max(0,\gamma+Y_{hrt}\phi(h,r,t)) \nonumber \\
&+ \lambda_1 \|Q\|^2 + \lambda_2\|W\|^2
\end{align}
where $\lambda_1$ and $\lambda_2$ are regularization coefficients controlling the strength of the $l_2$norm penalty.

To ensure efficient training and convergence, we adopt the initialization algorithm tailored for quaternion-valued networks as proposed by Parcollet et al\cite{parcollet2018quaternion}, building on principles from Glorot and Bengio \cite{glorot2010understanding} and He et al.\cite{he2015delving}. The initialization of entity and relation embeddings is defined as follows:
\begin{align}
   & w_{\rm real} = \rho \cos{(\theta)} , w_{{\rm img}i} = \rho \mathbf{Q}^{\triangleleft}_{{\rm img}i}\sin{(\theta)} \nonumber \\
   & w_{{\rm img}j} = \rho \mathbf{Q}^{\triangleleft}_{{\rm img}j}\sin{(\theta)} , w_{{\rm img}k} = \rho \mathbf{Q}^{\triangleleft}_{{\rm img}k}\sin{(\theta)},
\end{align}
where $ w_{\rm real},w_i,wj,w_k$ represent the scalar and imaginary parts of the quaternion embeddings. The parameter $\theta$  is uniformly sampled from $[ -\pi, \pi]$, and $ \mathbf{Q}^{\triangleleft}_{\rm img} $ is a normalized quaternion with a zero scalar part. The scaling factor $ \rho$  is randomly selected from the interval $ [\frac{-1}{\sqrt{2k}},\frac{1}{\sqrt{2k}}]$, where $k$ denotes the embedding dimensionality. This initialization strategy balances the parameter scales, accelerates model convergence, and enhances overall training efficiency.
\section{Theoretical Properties of QuatE-D}
Our quaternion-based embedding model provides significant advantages for knowledge graph representation. By leveraging quaternion algebra, the model effectively encodes the inherent properties of real-world relationships. 
\subsection{Non-Commutativity and Associativity}
Quaternions possess two key properties—non-commutativity and associativity—that are crucial for capturing the directionality and hierarchical nature of relationships. Non-commutative multiplication allows the model to encode directional relationships accurately, which is essential for asymmetric interactions. For instance, in a university setting, if Professor $A$ teaches subject $B$($r_1$), and Subject $B$ is a prerequisite for Subject $C$($r_2$)
\begin{itemize}
    \item $A \overset{r_1: {\rm teaches}}{\longrightarrow}B$
    \item $B \overset{r_2: {\rm prerequisite for}}{\longrightarrow}C$.
\end{itemize}

The composition $r_1 \circ r_2$  implies that $C$  is indirectly facilitated by $A$. Conversely, reversing the order($r_2 \circ r_1$) might suggest that $A$ is teaching a course required by $B$, a different relational interpretation. Non-commutativity ensures these distinctions are faithfully preserved in the embeddings.

Associativity enables consistent reasoning over chains of relationships, which is particularly useful for modeling hierarchical or multi-step dependencies. Consider the relationships in a university
\begin{itemize}
    \item $P \overset{r_1: {\rm supervises}}{\longrightarrow}Q$
    \item $Q \overset{r_2: {\rm advises}}{\longrightarrow} R$
    \item $R \overset{r_3:{\rm mentors}}{\longrightarrow} S $.
\end{itemize}
Using associativity, $( r_1 \circ r_2 )\circ r_3 = r_1 \circ (r_2 \circ r_3)$ , ensuring consistent inference that $P$ indirectly mentors $S$ . This property guarantees reliable reasoning over hierarchical relationships.
\subsection{Relational  Modeling}
The ability to accurately represent relational patterns is a hallmark of effective knowledge graph embeddings. Our quaternion-based approach excels in encoding the following patterns:
\subsubsection{Symmetry and Antisymmetry}
Symmetry captures bidirectional relationships$(r(x,y) \rightarrow r(y,x)$, such as two researchers collaborating $ A \overset{r}{\leftrightarrow} B )$, while antisymmetry represents one-directional dependencies$(r(x,y) \rightarrow  \neg r(y,x) ) $. The embeddings ensure that these relational nuances are preserved,
by setting the imaginary parts of $\mathbf{W}$ to zero for symmetry relations.


If the relation $r(x,y)$  holds true and $r(y,x)$ also holds, then the embeddings satisfy:
\begin{align}
     y = r \circ x  \land x = r \circ y  \rightarrow r \circ r = 1
\end{align}
This implies that ensuring the relation is symmetric. This property can be proven by constraining the imaginary parts of the quaternion $r$ to zero, effectively reducing $r$ to a real scalar, which enforces symmetry.

On the other hand, if $r(x,y)$  holds but $r(y,x)$ does not
\begin{align}
     y = r \circ x  \land  x \neq  r \circ y  \rightarrow r \circ r \neq 1
\end{align}
This condition ensures the relation is not symmetric. 
To establish the antisymmetry property, we focus on scenarios where the imaginary components of the quaternion $r$ are non-zero. Under this condition, the following inequality must hold:
\begin{align}\label{anty}
   \|{\mathbf{Q}}'_h-{\mathbf{Q}}_t\|^2 
   \neq 
   \|\mathbf{Q}'_t-\mathbf{Q}_h\|^2 
\end{align}
where ${\mathbf{Q}}'_h = \mathbf{Q}_h\otimes \mathbf{W}_r^{\triangleleft}$ and ${\mathbf{Q}}'_t = \mathbf{Q}_t\otimes \mathbf{W}_r^{\triangleleft}$. 
Expanding both sides of inequality \eqref{anty} reveals the terms:
\begin{align}\label{right}
&\|\mathbf{Q}'_h-\mathbf{Q}_t\|^2 = \|\mathbf{Q}'_h\|^2  + \|\mathbf{Q}_t\|^2 -2<\mathbf{Q}'_h,\mathbf{Q}_t>\nonumber \\
& \stackrel{|A\otimes B| = |A| \times |B|}{=} \|\mathbf{Q}_h\|^2\|\mathbf{W}_r^{\triangleleft}\|^2 +  \|\mathbf{Q}_t\|^2 -2<\mathbf{Q}'_h,\mathbf{Q}_t> \\ 
 & \|\mathbf{Q}'_t-\mathbf{Q}_h\|^2 =   \|\mathbf{Q}'_t\|^2  + \|\mathbf{Q}_h\|^2 -2<\mathbf{Q}'_t,\mathbf{Q}_h>\nonumber \\
 &   \stackrel{|A\otimes B| = |A| \times |B|}{=}  \|\mathbf{Q}_t\|^2\|\mathbf{W}_r^{\triangleleft}\|^2 +  \|\mathbf{Q}_h\|^2 -2<\mathbf{Q}'_t,\mathbf{Q}_h>
\end{align}
for the right and left-hand side. From these expressions, it is evident that proving antisymmetry requires showing that the inner products $<\mathbf{Q}'_h,\mathbf{Q}_t>  \neq <\mathbf{Q}'_t,\mathbf{Q}_h> $.These inner products can be expanded as follows:
\begin{align} \label{Q'hQt}
&<\mathbf{Q}'_h,\mathbf{Q}_t> \stackrel{\eqref{hamiltonKGE}}{=} \nonumber \\
& [(\mathbf{a}_h \circ \mathbf{p} - \mathbf{b}_h \circ \mathbf{q} - \mathbf{c}_h \circ \mathbf{u} - \mathbf{d}_h \circ \mathbf{v}) \nonumber \\
&+(\mathbf{a}_h \circ \mathbf{q} +\mathbf{b}_h \circ \mathbf{p} + \mathbf{c}_h \circ \mathbf{v}- \mathbf{d}_h \circ \mathbf{u})i\nonumber \\
&+ (\mathbf{a}_h \circ \mathbf{u}- \mathbf{b}_h \circ \mathbf{v} + \mathbf{c}_h \circ \mathbf{p} + \mathbf{d}_h \circ \mathbf{q})j\nonumber\\
&+(\mathbf{a}_h \circ \mathbf{v}+ \mathbf{b}_h \circ \mathbf{u} - \mathbf{c}_h \circ \mathbf{q} + \mathbf{d}_h \circ \mathbf{p})k] \nonumber \\
&\circ (\mathbf{a}_t + \mathbf{b}_ti + \mathbf{c}_tj + \mathbf{d}_tk) 
    \nonumber \\
& = (\mathbf{a}_h \circ \mathbf{p} - \mathbf{b}_h \circ \mathbf{q} - \mathbf{c}_h \circ \mathbf{u} - \mathbf{d}_h \circ \mathbf{v})\circ \mathbf{a}_t \nonumber \\
& +(\mathbf{a}_h \circ \mathbf{q} +\mathbf{b}_h \circ \mathbf{p} + \mathbf{c}_h \circ \mathbf{v}- \mathbf{d}_h \circ \mathbf{u}) \circ
    \mathbf{b}_t
\nonumber \\
&+ (\mathbf{a}_h \circ \mathbf{u}- \mathbf{b}_h \circ \mathbf{v} + \mathbf{c}_h \circ \mathbf{p} + \mathbf{d}_h \circ \mathbf{q}) \circ \mathbf{c}_t
\nonumber \\
& +(\mathbf{a}_h \circ \mathbf{v}+ \mathbf{b}_h \circ \mathbf{u} - \mathbf{c}_h \circ \mathbf{q} + \mathbf{d}_h \circ \mathbf{p}) \circ \mathbf{d}_t \nonumber \\
&=  <\mathbf{a}_h , \mathbf{p} ,\mathbf{a}_t> -   <\mathbf{b}_h , \mathbf{q} ,\mathbf{a}_t> -  <\mathbf{c}_h , \mathbf{u} ,\mathbf{a}_t> 
\nonumber \\
& - <\mathbf{d}_h , \mathbf{v} ,\mathbf{a}_t>  +<\mathbf{a}_h , \mathbf{q} ,\mathbf{b}_t> +   <\mathbf{b}_h , \mathbf{p} ,\mathbf{b}_t> \nonumber \\
& + <\mathbf{c}_h , \mathbf{v} ,\mathbf{b}_t> -   <\mathbf{d}_h , \mathbf{u} ,\mathbf{b}_t>
+ <\mathbf{a}_h , \mathbf{u} ,\mathbf{c}_t> \nonumber \\
&-   <\mathbf{b}_h , \mathbf{v} ,\mathbf{c}_t> -  <\mathbf{c}_h , \mathbf{p} ,\mathbf{c}_t> -   <\mathbf{d}_h , \mathbf{q} ,\mathbf{c}_t>
\nonumber \\
&+<\mathbf{a}_h , \mathbf{v} ,\mathbf{d}_t> +   <\mathbf{b}_h , \mathbf{u} ,\mathbf{d}_t> + <\mathbf{c}_h , \mathbf{q} ,\mathbf{d}_t>\nonumber \\
&-   <\mathbf{d}_h , \mathbf{p} ,\mathbf{d}_t>
\end{align}
and 
\begin{align} 
&<\mathbf{Q}'_t,\mathbf{Q}_h> \stackrel{\eqref{hamiltonKGE}}{=} \nonumber \\
    & [(\mathbf{a}_t \circ \mathbf{p} - \mathbf{b}_t \circ \mathbf{q} - \mathbf{c}_t \circ \mathbf{u} - \mathbf{d}_t \circ \mathbf{v}) \nonumber \\
    &+(\mathbf{a}_t \circ \mathbf{q} +\mathbf{b}_t \circ \mathbf{p} + \mathbf{c}_h \circ \mathbf{v}- \mathbf{d}_h \circ \mathbf{u})i\nonumber \\
    &+ (\mathbf{a}_t \circ \mathbf{u}- \mathbf{b}_t \circ \mathbf{v} + \mathbf{c}_t \circ \mathbf{p} + \mathbf{d}_t \circ \mathbf{q})j\nonumber\\
    &+(\mathbf{a}_t \circ \mathbf{v}+ \mathbf{b}_t \circ \mathbf{u} - \mathbf{c}_t \circ \mathbf{q} + \mathbf{d}_t \circ \mathbf{p})k] \nonumber \\
    &\circ (\mathbf{a}_h + \mathbf{b}_hi + \mathbf{c}_hj + \mathbf{d}_hk) 
    \nonumber \\
    & = (\mathbf{a}_t \circ \mathbf{p} - \mathbf{b}_t \circ \mathbf{q} - \mathbf{c}_t \circ \mathbf{u} - \mathbf{d}_t \circ \mathbf{v})\circ \mathbf{a}_h \nonumber \\
    & +(\mathbf{a}_t \circ \mathbf{q} +\mathbf{b}_t \circ \mathbf{p} + \mathbf{c}_t \circ \mathbf{v}- \mathbf{d}_t\circ \mathbf{u}) \circ
    \mathbf{b}_h
\nonumber \\
&+ (\mathbf{a}_t \circ \mathbf{u}- \mathbf{b}_t \circ \mathbf{v} + \mathbf{c}_t \circ \mathbf{p} + \mathbf{d}_t \circ \mathbf{q}) \circ \mathbf{c}_h
\nonumber \\
& +(\mathbf{a}_t \circ \mathbf{v}+ \mathbf{b}_t \circ \mathbf{u} - \mathbf{c}_t \circ \mathbf{q} + \mathbf{d}_t \circ \mathbf{p}) \circ \mathbf{d}_h \nonumber \\
&=  <\mathbf{a}_t , \mathbf{p} ,\mathbf{a}_h> -   <\mathbf{b}_t , \mathbf{q} ,\mathbf{a}_h> -  <\mathbf{c}_t , \mathbf{u} ,\mathbf{a}_h> 
\nonumber \\
& - <\mathbf{d}_t , \mathbf{v} ,\mathbf{a}_h>  +<\mathbf{a}_t , \mathbf{q} ,\mathbf{b}_h> +   <\mathbf{b}_t , \mathbf{p} ,\mathbf{b}_h> \nonumber \\
& + <\mathbf{c}_t , \mathbf{v} ,\mathbf{b}_h> -   <\mathbf{d}_t , \mathbf{u} ,\mathbf{b}_h>
+ <\mathbf{a}_t , \mathbf{u} ,\mathbf{c}_h> \nonumber \\
&-   <\mathbf{b}_t , \mathbf{v} ,\mathbf{c}_h> -  <\mathbf{c}_t , \mathbf{p} ,\mathbf{c}_h> -   <\mathbf{d}_t , \mathbf{q} ,\mathbf{c}_h>
\nonumber \\
&+<\mathbf{a}_t , \mathbf{v} ,\mathbf{d}_h> +   <\mathbf{b}_t , \mathbf{u} ,\mathbf{d}_h> + <\mathbf{c}_t , \mathbf{q} ,\mathbf{d}_h>\nonumber \\
&-   <\mathbf{d}_t , \mathbf{p} ,\mathbf{d}_h>
\end{align}
Due to the non-commutative nature of quaternion multiplication, these inner products differ, ensuring that:
\begin{align}
    <\mathbf{Q}'_h,\mathbf{Q}_t>  \neq <\mathbf{Q}'_t,\mathbf{Q}_h> 
\end{align}
Thus, the antisymmetry property is satisfied when the imaginary components of $r$ are non-zero.
\subsubsection{Inversion}
Inversion enables the model to represent reverse relationships.For instance, if Professor $P$  supervises Student $S$ ($P \overset{r_1}{\rightarrow}S$), the inverse relation $S \overset{r_2}{\rightarrow}P$  correctly reflects that $S$  is supervised by $P$
,where $r_2$ inverse of $r_1$.

To demonstrate the property of inversion, consider a pair of relations$(r_1(x,y) \rightarrow r_2(y,x)$, if $ y = r_1 \circ x  \land x = r_2\circ y  $, it follows that $r_1 = r_2^{-1}$.
This result utilizes the conjugate of quaternions, which serves as the inverse operation under quaternion algebra. To verify this property, we examine the equality:
\begin{align}\label{inverse}
     \|\mathbf{Q}'_h-\mathbf{Q}_t\|^2 = \|\mathbf{Q}'_t-\mathbf{Q}_h\|^2.
\end{align}
Expanding the left and right-hand side of \eqref{inverse}, we get:
\begin{align}\label{right&left}
     &\|\mathbf{Q}'_h-\mathbf{Q}_t\|^2 = \|\mathbf{Q}'_h\|^2  + \|\mathbf{Q}_t\|^2 -2<\mathbf{Q}'_h,\mathbf{Q}_t>\nonumber \\
     & \stackrel{|A\otimes B| = |A| \times |B|}{=} \|\mathbf{Q}_h\|^2\|\mathbf{W}_r^{\triangleleft}\|^2 +  \|\mathbf{Q}_t\|^2 -2<\mathbf{Q}'_h,\mathbf{Q}_t> \\ 
     & \|\mathbf{Q}'_t-\mathbf{Q}_h\|^2 =   \|\mathbf{Q}'_t\|^2  + \|\mathbf{Q}_h\|^2 -2<\mathbf{Q}'_t,\mathbf{Q}_h>\nonumber \\
 &   \stackrel{|A\otimes B| = |A| \times |B|}{=}  \|\mathbf{Q}_t\|^2\|\overline{\mathbf{W}}_r^{\triangleleft}\|^2 +  \|\mathbf{Q}_h\|^2 -2<\mathbf{Q}'_t,\mathbf{Q}_h>
\end{align}
By the property of quaternion conjugation, $|\mathbf{W}_r^{\triangleleft}\|=|\overline{\mathbf{W}}_r^{\triangleleft}\|=1$, it follows that 
\begin{align}
    \|\mathbf{Q}^{'}_h\|^2 + \|\mathbf{Q}_t\|^2 = \|\mathbf{Q}'_t\|^2 + \|\mathbf{Q}_h\|^2.
\end{align}
Additionally, the equality of the 
\begin{align}
    <\mathbf{Q}'_h,\mathbf{Q}_t> =   <\mathbf{Q}'_t,\mathbf{Q}_h> 
\end{align}
is verified explicitly (\eqref{Q'hQt} and \eqref{Q'tQhinverse}).
\begin{align} \label{Q'tQhinverse}
&<\mathbf{Q}'_t,\mathbf{Q}_h>  \stackrel{\eqref{hamiltonKGE}}{=} \nonumber \\
    & [(\mathbf{a}_t \circ \mathbf{p} +\mathbf{b}_t \circ \mathbf{q} + \mathbf{c}_t \circ \mathbf{u} + \mathbf{d}_t \circ \mathbf{v}) \nonumber \\
    &+(-\mathbf{a}_t \circ \mathbf{q} +\mathbf{b}_t \circ \mathbf{p} - \mathbf{c}_h \circ \mathbf{v}+\mathbf{d}_h \circ \mathbf{u})i\nonumber \\
    &+ (-\mathbf{a}_t \circ \mathbf{u}+ \mathbf{b}_t \circ \mathbf{v} + \mathbf{c}_t \circ \mathbf{p} - \mathbf{d}_t \circ \mathbf{q})j\nonumber\\
    &+(-\mathbf{a}_t \circ \mathbf{v}- \mathbf{b}_t \circ \mathbf{u} +\mathbf{c}_t \circ \mathbf{q} + \mathbf{d}_t \circ \mathbf{p})k] \nonumber \\
    &\circ (\mathbf{a}_h + \mathbf{b}_hi + \mathbf{c}_hj + \mathbf{d}_hk) 
    \nonumber \\
    & = (\mathbf{a}_t \circ \mathbf{p} +\mathbf{b}_t \circ \mathbf{q} + \mathbf{c}_t \circ \mathbf{u} + \mathbf{d}_t \circ \mathbf{v})\circ \mathbf{a}_h \nonumber \\
    & +(-\mathbf{a}_t \circ \mathbf{q} +\mathbf{b}_t \circ \mathbf{p} - \mathbf{c}_h \circ \mathbf{v}+\mathbf{d}_h \circ \mathbf{u}) \circ
    \mathbf{b}_h
\nonumber \\
&+ (-\mathbf{a}_t \circ \mathbf{u}+ \mathbf{b}_t \circ \mathbf{v} + \mathbf{c}_t \circ \mathbf{p} - \mathbf{d}_t \circ \mathbf{q}) \circ \mathbf{c}_h
\nonumber \\
& +(-\mathbf{a}_t \circ \mathbf{v}- \mathbf{b}_t \circ \mathbf{u} +\mathbf{c}_t \circ \mathbf{q} + \mathbf{d}_t \circ \mathbf{p}) \circ \mathbf{d}_h \nonumber \\
&=  <\mathbf{a}_t , \mathbf{p} ,\mathbf{a}_h> + <\mathbf{b}_t , \mathbf{q} ,\mathbf{a}_h> +  <\mathbf{c}_t , \mathbf{u} ,\mathbf{a}_h> 
\nonumber \\
& +<\mathbf{d}_t , \mathbf{v} ,\mathbf{a}_h>  -<\mathbf{a}_t , \mathbf{q} ,\mathbf{b}_h> +   <\mathbf{b}_t , \mathbf{p} ,\mathbf{b}_h> \nonumber \\
& - <\mathbf{c}_t , \mathbf{v} ,\mathbf{b}_h> +   <\mathbf{d}_t , \mathbf{u} ,\mathbf{b}_h>
- <\mathbf{a}_t , \mathbf{u} ,\mathbf{c}_h> \nonumber \\
&+   <\mathbf{b}_t , \mathbf{v} ,\mathbf{c}_h> + <\mathbf{c}_t , \mathbf{p} ,\mathbf{c}_h> -   <\mathbf{d}_t , \mathbf{q} ,\mathbf{c}_h>
\nonumber \\
&-<\mathbf{a}_t , \mathbf{v} ,\mathbf{d}_h> +   <\mathbf{b}_t , \mathbf{u} ,\mathbf{d}_h> + <\mathbf{c}_t , \mathbf{q} ,\mathbf{d}_h>\nonumber \\
&+   <\mathbf{d}_t , \mathbf{p} ,\mathbf{d}_h>.
\end{align}
Now for  $ \|\mathbf{Q}^{'}_h\|^2 + \|\mathbf{Q}_t\|^2 = \|\mathbf{Q}'_t\|^2 + \|\mathbf{Q}_h\|^2$, we utilize the property that $ |A\otimes B| = |A| \times |B| $ . Consequently, we can write 
\begin{align}
     &\|\mathbf{Q}'_h\|^2 + \|\mathbf{Q}_t\|^2 = \|\mathbf{Q}_h\|^2 |\mathbf{W}_r^{\triangleleft}\|^2+ \|\mathbf{Q}_t\|^2 \nonumber \\
     &=\|\mathbf{Q}_h\|^2 +  \|\mathbf{Q}_t\|^2 |\overline{\mathbf{W}}_r^{\triangleleft}\|^2 =\|\mathbf{Q}'_t\|^2 + \|\mathbf{Q}_h\|^2
\end{align}
because $|\mathbf{W}_r^{\triangleleft}\|=|\overline{\mathbf{W}}_r^{\triangleleft}\|=1$.
\subsubsection{Composition}
Composition models the combination of multiple relationships to infer new connections. For example, if department dead $H$ oversees Professor $P$($H \overset{r_1}{\rightarrow} P$) , and Professor $P$  advises Student $S$( $P \overset{r_2}{\rightarrow}S$) , the composition $r_1 \circ r_2$ infers that $H$  indirectly advises $S$.

If $r_1(x,z),r_2(x,y)$ and $r_3(y,z)$ hold
\begin{align}
    z=r_1\circ y \land y = r_3\circ x \land x = r_2 \circ y \rightarrow r_1 = r_2 \circ r_3.
\end{align}
Using the quaternion embedding framework, the composition embedding can be expressed as \eqref{comp} by the associativity of quaternion multiplication.
\begin{align}\label{comp}
    &\|(\mathbf{Q}_h \otimes \mathbf{W}^{\triangleleft}_{r2})\otimes \mathbf{W}^{\triangleleft}_{r3} -\mathbf{Q}_t\|_2 \nonumber \\
    &= \|\mathbf{Q}_h \otimes (\mathbf{W}^{\triangleleft}_{r2}\otimes \mathbf{W}^{\triangleleft}_{r3}) -\mathbf{Q}_t\|_2 \nonumber \\
    & =\|\mathbf{Q}_h \otimes \mathbf{W}^{\triangleleft}_{r1} -\mathbf{Q}_t\|_2.
\end{align} 
\subsection{Reduction of QUATE-D}
Our proposed model, QUATE-D, utilizes quaternion embeddings and the Hamilton product in the scoring function \eqref{score}. This scoring function demonstrates flexibility by naturally reducing to specific existing models under certain constraints on the quaternion embeddings. When quaternion embeddings are constrained to the complex plane ($\mathbf{Q}_h=\mathbf{a}_h+\mathbf{b}_hi, \mathbf{W}_h=\mathbf{a}_r+\mathbf{b}_ri, \mathbf{Q}_t=\mathbf{a}_t+\mathbf{b}_ti$), the Hamilton product simplifies to complex multiplication:
\begin{align}
     &\mathbf{Q}'_h(\mathbf{a}'_h,\mathbf{b}'_h,\mathbf{c}'_h,\mathbf{d}'_h) = \mathbf{Q}_h \otimes  \mathbf{W}_r^{\triangleleft}=\nonumber \\
    & (\mathbf{a}_h \circ \mathbf{p} - \mathbf{b}_h \circ \mathbf{q})+     (\mathbf{a}_h \circ \mathbf{q} +\mathbf{b}_h \circ \mathbf{p} )i
\end{align}
Substituting into the scoring function, we obtain:
\begin{align}
    \phi(h,r,t)=\|\mathbf{Q}'_h-\mathbf{Q}_t\|_2= 
=\sqrt{(a'_h-a_t)^2+(b'_h-b_t)^2}.
\end{align}
 This is equivalent to the scoring function of RotatE\cite{sun2019rotate}
\section{Experiments and Results}
\subsection{Dataset}
The evaluation of knowledge graph embedding models commonly relies on well-established benchmark datasets. Among them, WN18\cite{bordes2013translating}  and FB15k\cite{bordes2013translating} are particularly notable. WN18 is a subset of WordNet\cite{miller1995wordnet}, a lexical database of semantic relationships between words. This dataset includes 40,943 entities and 18 relations, forming a total of 151,442 triples. The relations in WN18 capture lexical relationships such as synonymy and hypernymy. FB15k, derived from the Freebase knowledge graph\cite{bollacker2008freebase}, is a collection of general-purpose facts. It contains 14,951 entities and 1,345 relations, leading to a total of 592,213 triples. These relations encompass diverse domains such as people, locations, and films.

Despite their widespread use, these datasets have been criticized for their reliance on inverse relations, which allow trivial inference of test triples by reversing training triples. To mitigate this, refined datasets—WN18RR and FB15k-237—were introduced:WN18RR \cite{dettmers2018convolutional} modifies WN18 by removing redundant inverse relations, leaving 11 relations and a total of 93,003 triples.FB15k-237\cite{toutanova2015observed} improves FB15k by reducing redundancy and limiting the number of relations to 237, resulting in a dataset of 310,116 triples.

These enhanced versions offer more robust benchmarks to evaluate the generalization capabilities of knowledge graph embedding models. The statistics of these datasets are summarized in \ref{tabent}.
\begin{table}[t]
    \centering
    \caption{Key statistics of the WN18, FB15k, WN18RR, and FB15k-237 datasets.}
    \resizebox{0.45\textwidth}{!}{\begin{tabular}{|c|c|c|c|c|c|c|}
    \hline 
         Dataset	&Entities	&Relations	&Triples&	Train	&Validation	& Test\\
        \hline
        WN18 &	40,943	&18&	151,442	&141,442 &	5,000 	& 5,000 \\
        \hline 
        FB15k	 &14,951&	1,345	&592,213	&483,142&	50,000	&59,071\\
        \hline 
WN18RR &	40,943	&11	&93,003&	86,835	&3,034	&3,134 \\
\hline
FB15k-237	&14,541 &237	&310,116	&272,115	&17,535&	20,466\\
\hline
    \end{tabular}}
    \label{tabent}
\end{table}
\subsection{Evaluation Protocol}
To evaluate the performance of our model, we adopt three widely used metrics: Mean Rank (MR), Mean Reciprocal Rank (MRR), and Hits@$n$ with cut-off values 
$n=1,3,10$. MR calculates the average rank of all correct entities, with lower values indicating better performance. MRR reflects the average of the reciprocal ranks for correct entities, providing a more fine-grained evaluation. Hits@$n$ measures the proportion of correct entities that appear within the top $n$ predictions. Following the protocol established by \cite{bordes2013translating}, we report filtered results to mitigate the impact of potentially flawed evaluations due to the presence of other valid entities in the ranking.
\subsection{Baseline}
To evaluate the effectiveness of our model, we compare it against several established baseline methods, including TransE\cite{bordes2013translating}, RotatE\cite{sun2019rotate}, ComplEx\cite{trouillon2016complex}, DistMult\cite{yang2014embedding}, ConvE\cite{dettmers2018convolutional}, R-GCN\cite{schlichtkrull2018modeling}, NKGE\cite{wang2018knowledge}, HolE\cite{nickel2016holographic}, QuatE\cite{zhang2019quaternion}, and DualE\cite{cao2021dual}. These models represent a diverse set of approaches to knowledge graph embedding, ranging from translational models to semantic matching and neural architectures.
\subsection{Implementation Details}
Our model was implemented using PyTorch and tested on a single GPU for all experiments. Hyperparameters were determined via grid search, and the optimal models were selected using early stopping based on validation set performance. The embedding size $k$ was chosen from $\{50,100,150,200,300,400\}$, while the regularization rates $\lambda_1$ and $\lambda_2$  were tuned from $\{0,0.05,1\}$
.The learning rate ($\alpha$) for Adagrad \cite{duchi2011adaptive} optimization
 was selected $\{0.02\}$, and the negative sampling rate (\#neg) was chosen from $\{1,5,10\}$. A fixed margin $\gamma=1$ was used throughout all experiments.

The batch size was fixed at 10 for all datasets, ensuring consistency across our model (QuatE-D), QuatE\cite{zhang2019quaternion}, and RotatE\cite{sun2019rotate}. This choice allowed for efficient training while keeping the computational load manageable. The number of epochs required for training each model is provided in Table 6 for QuatE-D, QuatE, and RotatE. These values were determined through early stopping, ensuring that each model was trained for an adequate number of epochs to reach convergence without overfitting.
\begin{table}[t]
    \centering
    \caption{The number of epochs need of QuatE-D, QuatE, and RotatE}
    \resizebox{0.45\textwidth}{!}{\begin{tabular}{|c|c|c|c|c|}
    \hline 
         Datasets	& WN18	& WN18RR	& FB15k &	FB15k-237	\\
        \hline
        QuatE-D &	3000 & 30000 & 5000 & 5000 \\
        \hline 
        QuatE	 &3000 & 30000 & 5000 & 5000 \\
        \hline 
    RotatE &	80000	& 80000	& 150000 & 150000 \\
\hline
    \end{tabular}}
    \label{tableepoch}
\end{table}
\subsection{Results}
Tables \ref{tabcompareWN18&FB15k} and \ref{tabcompareWN18rr&FB15k237} present the link prediction results on four benchmark datasets, evaluating various knowledge graph embedding models.

In Table \ref{tabcompareWN18&FB15k}, for WN18 QuatE-D$^2$  achieves the best MR $(160)$, significantly improving upon previous methods, including RotatE $({\rm MR} = 184)$.  In terms of MRR, QuatE-D$^2$ achieves $0.948$, which is competitive with DuatE $(0.951)$ and QuatE $(0.949)$. QuatE-D$^2$ also attains the best Hit@3 $(0.956)$ and second-best Hit@1 $(0.942)$, demonstrating its robustness.

On FB15K, QuatE-D$^2$ outperforms other models in several metrics, achieving the lowest MR $(21)$ and the best Hit@3 $(0.831)$ and MRR $ (0.786)$. Additionally, it achieves the highest Hit@10 $(0.892)$, matching DistMult, and secures the second-best Hit@1 $(0.722)$. Notably, while DuatE achieves the highest MRR $(0.790)$ and Hit@1 $(0.734)$, it benefits from greater degrees of freedom compared to QuatE and QuatE-D.

While our model does not achieve the best results across all metrics, it still performs competitively, particularly in reducing Mean Rank. It is important to note that DuatE benefits from a higher degree of freedom, as it effectively utilizes twice the number of parameters compared to our model. This additional capacity allows DuatE to capture more complex relationships, which explains its superior MRR and Hit@K scores. Nevertheless, our quaternion-based approach remains effective and demonstrates strong generalization across datasets.

In Table \ref{tabcompareWN18rr&FB15k237}, on the more challenging WN18RR dataset, QuatE-D$^2$ achieves the best MR $(1050)$, outperforming RotatE $(3277) $and QuatE $(3472)$. It also attains the highest MRR $(0.483)$, surpassing DuatE $(0.482)$ and QuatE $(0.481)$. In terms of Hit@1, QuatE-D achieves $0.447$, the best among all models, while matching RotatE in Hit@10 $(0.565)$.

For FB15K-237, QuatE-D$^2$ consistently outperforms existing approaches, achieving the lowest MR $(71)$ and the highest MRR $(0.443)$. Additionally, it achieves the best Hit@10 $(0.638)$, Hit@3 $(0.487)$, and Hit@1 $(0.344)$, outperforming all baseline methods, including DuatE and RotatE.

QuatE-D consistently demonstrates strong performance across all datasets, outperforming most baselines and achieving state-of-the-art results in several metrics. The improvements in MR, MRR, and Hit@K suggest that the distance-based scoring function enhances the model’s ability to capture relational dependencies effectively. Compared to DuatE, which benefits from a higher degree of freedom, QuatE-D still achieves competitive performance, especially on WN18RR and FB15K-237. Furthermore, our experiments confirm that DistMult exhibits higher negative sampling rates, which may affect its overall ranking performance.

Additionally, both QuatE and DuatE were optimized under more extensive hyperparameter tuning, while QuatE-D achieves superior results under limited hyperparameter settings. This indicates that further optimization could enhance the performance of our model even more. Overall, the results validate the effectiveness of our proposed quaternion-based approach in knowledge graph embedding.

\begin{figure}
    \centering
    \includegraphics[width=\linewidth]{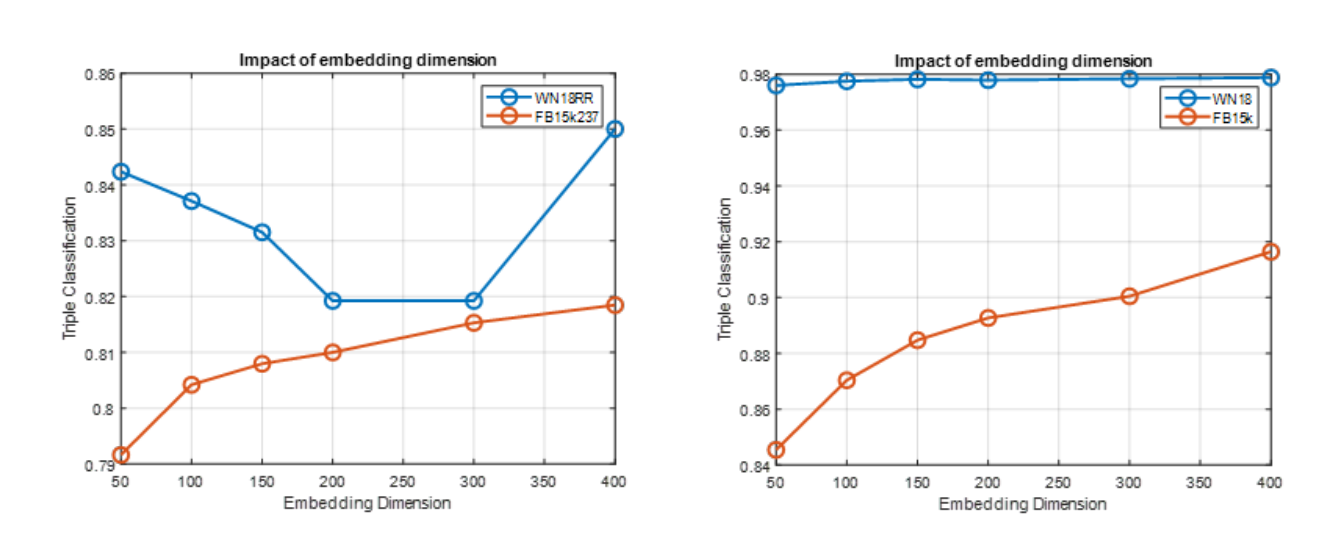}
    \caption{Impact of Embedding Dimension on Triple Classification Performance}
    \label{fig1}
\end{figure}

\begin{table*}[t]
    \centering
    \caption{
    Link prediction results on WN18 and FB15K. The best results are in bold and the second best results are underlined.$[\circ]$: 
    Results are taken from\cite{nickel2016holographic}; 
    $[\lozenge]$: Results are taken from \cite{dettmers2018convolutional}; 
    Other results are taken from the corresponding original papers.
    [QuatE-D $^1$] : without type constraints; [QuatE-D $^2$] :  with type constraints.}
\begin{tabular}{|c|c|c|c|c|c|c|c|c|c|c|}
\hline &  \multicolumn{5}{c}{ WN18 } &  \multicolumn{5}{|c|}{ FB15K } \\
\cline{2-11}
Model & MR & MRR & Hit@10 & Hit@3 & Hit@1 & MR & MRR & Hit@10 & Hit@3 & Hit@1 \\
\hline TransE$^\circ$ & - & 0.495 & 0.943 & 0.888 & 0.113 & - & 0.463 & 0.749 & 0.578 & 0.297 \\
DistMult$^\lozenge$ & 655 & 0.797 & 0.946 & - & - & 42.2 & 0.798 & $\mathbf{0.893}$ & - & - \\
ComplEx& - & 0.941 & 0.947 & 0.945 & 0.936 & - & 0.692 & 0.840 & 0.759 & 0.599 \\
ConvE& 374 & 0.943 & 0.956 & 0.946 & 0.935 & 51 & 0.657 & 0.831 & 0.723 & 0.558 \\
HolE & - & 0.938 & 0.949 & 0.945 & 0.930 & - & 0.524 & 0.739 & 0.759 & 0.599 \\
R-GCN+ & - & 0.819 & $\mathbf{0 . 9 6 4}$ & 0.929 & 0.697 & - & 0.696 & 0.842 & 0.760 & 0.601 \\
NKGE & 336 & 0.947 & 0.957 & 0.949 & 0.942 & 56 & 0.73 & 0.871 & 0.790 & 0.650 \\
RotatE & $\underline{184}$ & 0.947 & 0.961 & $0.953$ & 0.938 & 40 & 0.699 & 0.872 & 0.788& 0.585\\
 QuatE  & 388 & $\underline{0 . 9 4 9}$ & 0.960 & $\underline{0 . 9 5 4}$ & 0.941 & $\underline{41}$ & 0.770 & 0.878 & 0.821 & 0.700 \\
DuatE  & - & $\mathbf{0 . 9 5 1}$ & $\underline{0.961}$ & $\mathbf{0 . 9 5 6}$ & $\mathbf{0 . 9 4 5}$ & - & $\mathbf{0 . 790}$ & 0 . 881 & $\underline{0 . 829}$ & $\mathbf{0 . 734}$  \\
\hline
QuatE-D$^1$  & 366 & 0.946 & 0.955 & 0.950 & 0.941& 45 & 0.763 & 0.871 & 0.808 & 0.697 \\
QuatE-D$^2$   & $\mathbf{160}$ & 0.948 & 0.960 &$\mathbf{ 0.956}$& $\underline{0.942 }$& $\mathbf{21}$& $\underline{0.786}$ & $\underline{0.892}$& $\mathbf{0.831}$& $\underline{0.722}$ \\
\hline
\end{tabular}
    \label{tabcompareWN18&FB15k}
\end{table*}
\begin{table*}[t]
    \centering
    \caption{
    Link prediction results on WN18RR and FB15K-237. The best results are in bold and the second best results are underlined.$[\circ]$: 
    Results are taken from\cite{nguyen2017novel}; 
    $[\lozenge]$: Results are taken from \cite{dettmers2018convolutional}; $[\triangle]$: Results are taken from \cite{sun2019rotate}.
    Other results are taken from the corresponding original papers.
    QuatE-D $^{1}$: without type constraints; QuatE-D $^{2}$:  with type constraints.}
\begin{tabular}{|c|c|c|c|c|c|c|c|c|c|c|}
\hline &  \multicolumn{5}{c}{ WN18RR } &  \multicolumn{5}{|c|}{ FB15K237 } \\
\cline{2-11}
Model & MR & MRR & Hit@10 & Hit@3 & Hit@1 & MR & MRR & Hit@10 & Hit@3 & Hit@1 \\
\hline TransE$^\circ$ & 3384 &0.226 &0.501& - &- &357 &0.294 &0.465 &- &- \\
DistMult$^\lozenge$ &5110 &0.43 &0.49& 0.44& 0.39 &254& 0.241 &0.419 &0.263& 0.155 \\
ComplEx $^\lozenge$& 5261&0.44 &0.51 &0.46& 0.41& 339& 0.247 &0.428 &0.275 &0.158\\
ConvE ${^\lozenge}$& 4187 &0.43 &0.52 &0.44 &0.40 &244 &0.325& 0.501 &0.356 &0.237 \\
R-GCN+ & -& - &- &- &- &- &0.249 &0.417& 0.264 &0.151 \\
NKGE & 4170 &0.45 &0.526 &0.465 &0.421 &237 &0.33& 0.510 &0.365 &0.241 \\
RotatE$^\triangle$ & $\underline{3277}$& 0.470 & $\mathbf{0.565}$ &0.488 &0.422 &185& 0.297 & 0.480 & 0.328& 0.205 \\
 QuatE  & 3472& 0.481 & $\underline{0.564}$ & $\underline{0.500}$ &0.436& 176 &0.311 &0.495 &0.342 &0.221 \\
DuatE  & - & $\underline{0.482}$ &0.561 &$\underline{0.500}$  & $\underline{0.440}$ &- &0.330 &0.518 &0.363 &0.237 \\
\hline
QuatE-D$^1$  & 4206 & 0.468& 0.535 & 0.490 &0.436 & $\underline{131}$ & $\underline{0.425}$ & $\underline{0.617}$ & $\underline{0.469}$& $\underline{ 0.327}$ \\
QuatE-D$^2$   & $\mathbf{1050}$& $\mathbf{0.483 }$ &  $\mathbf{0.565}$ & $\mathbf{0.503}$ & $\mathbf{0.447}$ & $\mathbf{71}$ & $\mathbf{0.443 }$& $\mathbf{0.638}$ & $\mathbf{0.487}$ & $\mathbf{0.344}$  \\
\hline
\end{tabular}
    \label{tabcompareWN18rr&FB15k237}
\end{table*}
Table \ref{relationMRR} presents the Mean Reciprocal Rank (MRR) of different models on each relation within the WN18RR dataset. The results indicate that our proposed QuatE-D model performs best across most relations, highlighting its effectiveness in modeling complex relational structures.
\begin{table}[t]
    \centering
    \caption{ MRR for the models tested on each relation of WN18RR.}
    \resizebox{0.45\textwidth}{!}{\begin{tabular}{|c|c|c|c|}
\hline 
{\rm Relation Name} & {\rm RotatE} & {\rm QuatE}& \rm{Quat-D$^2$} \\
\hline hypernym & 0.148 & 0.173 & $\mathbf{ 0.207}$\\
\hline {\rm derivationally related form} & 0.947 & $\mathbf{0.953}$ & 0.952\\
\hline{\rm  instance hypernym} & 0.318 & 0.364 & $\mathbf{0.508}$\\
\hline {\rm also see} & 0.585 & 0.629& $\mathbf{0.694}$ \\
\hline {\rm member meronym} &0.232& 0.232 & $\mathbf{0.294}$ \\
\hline {\rm synset domain topic of} & 0.341 & 0.468 & $\mathbf{0.525}$ \\
\hline {\rm has part} & 0.184 & 0.233 & $\mathbf{0.251}$\\
\hline {\rm member of domain usage} & 0.318 &0.441 & $\mathbf{0.676}$ \\
\hline {\rm member of domain region} & 0.200 & 0.193 & $\mathbf{0.33}$ \\
\hline {\rm verb group} & 0.943 & 0.924 & $\mathbf{0.974}$ \\
\hline {\rm similar to} & $\mathbf{1 . 0 0 0}$ & $\mathbf{1 . 0 0 0}$ & $\mathbf{1 . 0 0 0}$\\
\hline
    \end{tabular}}
    \label{relationMRR}
\end{table}
Figure \ref{fig1} illustrates the effect of varying embedding dimensions on triple classification accuracy.  For WN18 and FB15K, performance improves with increasing embedding size, with WN18 maintaining consistently high accuracy across all dimensions, while FB15K shows a steady upward trend. In contrast, WN18RR demonstrates non-monotonic behavior, experiencing a decline at intermediate dimensions before improving at higher ones. FB15K-237 follows a trend similar to FB15K, with performance gradually increasing as the embedding dimension grows.

These results suggest that FB15K and FB15K-237 benefit from larger embeddings due to their complex relational structures, whereas WN18 remains stable across dimensions, requiring lower embedding sizes. The fluctuations in WN18RR indicate that embedding size has a non-linear effect, possibly due to overfitting or dataset-specific characteristics. Overall, higher embedding dimensions generally enhance performance, but dataset complexity and structure influence the optimal choice of dimensionality. 
\section{conclution}

In this paper, we introduced QuatE-D, a quaternion-based knowledge graph embedding model that utilizes a distance-based scoring function to improve relational modeling. Our results confirm that quaternion representations effectively capture structural dependencies while maintaining efficiency.

For future work, we aim to explore the integration of QuatE-D's scoring function into other quaternion-based models, such as DuatE\cite{cao2021dual} and QuatDE \cite{gao2021quatde}, to further assess its impact on different algebraic structures. Additionally, extending this approach to dynamic knowledge graphs and investigating its compatibility with other embedding techniques could enhance its adaptability and effectiveness in large-scale applications.
\section*{Acknowledgement}
We would like to express our sincere gratitude to Dr. Aliazam Abbassfar for his valuable feedback and insightful comments on this paper.

 
%

\bibliographystyle{ieeetr}
\bibliography{MyRefrences}

\begin{thebibliography}{10}

\bibitem{hao2017end}
Y.~Hao, Y.~Zhang, K.~Liu, S.~He, Z.~Liu, H.~Wu, and J.~Zhao, ``An end-to-end
  model for question answering over knowledge base with cross-attention
  combining global knowledge,'' in {\em Proceedings of the 55th Annual Meeting
  of the Association for Computational Linguistics (Volume 1: Long Papers)},
  pp.~221--231, 2017.

\bibitem{cui2019kbqa}
W.~Cui, Y.~Xiao, H.~Wang, Y.~Song, S.-w. Hwang, and W.~Wang, ``Kbqa: learning
  question answering over qa corpora and knowledge bases,'' {\em arXiv preprint
  arXiv:1903.02419}, 2019.

\bibitem{saxena2020improving}
A.~Saxena, A.~Tripathi, and P.~Talukdar, ``Improving multi-hop question
  answering over knowledge graphs using knowledge base embeddings,'' in {\em
  Proceedings of the 58th annual meeting of the association for computational
  linguistics}, pp.~4498--4507, 2020.

\bibitem{yang2019leveraging}
B.~Yang and T.~Mitchell, ``Leveraging knowledge bases in lstms for improving
  machine reading,'' {\em arXiv preprint arXiv:1902.09091}, 2019.

\bibitem{zhang2016collaborative}
F.~Zhang, N.~J. Yuan, D.~Lian, X.~Xie, and W.-Y. Ma, ``Collaborative knowledge
  base embedding for recommender systems,'' in {\em Proceedings of the 22nd ACM
  SIGKDD international conference on knowledge discovery and data mining},
  pp.~353--362, 2016.

\bibitem{wang2018dkn}
H.~Wang, F.~Zhang, X.~Xie, and M.~Guo, ``Dkn: Deep knowledge-aware network for
  news recommendation,'' in {\em Proceedings of the 2018 world wide web
  conference}, pp.~1835--1844, 2018.

\bibitem{bordes2013translating}
A.~Bordes, N.~Usunier, A.~Garcia-Duran, J.~Weston, and O.~Yakhnenko,
  ``Translating embeddings for modeling multi-relational data,'' {\em Advances
  in neural information processing systems}, vol.~26, 2013.

\bibitem{sun2019rotate}
Z.~Sun, Z.-H. Deng, J.-Y. Nie, and J.~Tang, ``Rotate: Knowledge graph embedding
  by relational rotation in complex space,'' {\em arXiv preprint
  arXiv:1902.10197}, 2019.

\bibitem{zhang2019quaternion}
S.~Zhang, Y.~Tay, L.~Yao, and Q.~Liu, ``Quaternion knowledge graph
  embeddings,'' {\em Advances in neural information processing systems},
  vol.~32, 2019.

\bibitem{bollacker2008freebase}
K.~Bollacker, C.~Evans, P.~Paritosh, T.~Sturge, and J.~Taylor, ``Freebase: a
  collaboratively created graph database for structuring human knowledge,'' in
  {\em Proceedings of the 2008 ACM SIGMOD international conference on
  Management of data}, pp.~1247--1250, 2008.

\bibitem{miller1995wordnet}
G.~A. Miller, ``Wordnet: a lexical database for english,'' {\em Communications
  of the ACM}, vol.~38, no.~11, pp.~39--41, 1995.

\bibitem{wang2014knowledge}
Z.~Wang, J.~Zhang, J.~Feng, and Z.~Chen, ``Knowledge graph embedding by
  translating on hyperplanes,'' in {\em Proceedings of the AAAI conference on
  artificial intelligence}, vol.~28, 2014.

\bibitem{lin2015learning}
Y.~Lin, Z.~Liu, M.~Sun, Y.~Liu, and X.~Zhu, ``Learning entity and relation
  embeddings for knowledge graph completion,'' in {\em Proceedings of the AAAI
  conference on artificial intelligence}, vol.~29, 2015.

\bibitem{ji2015knowledge}
G.~Ji, S.~He, L.~Xu, K.~Liu, and J.~Zhao, ``Knowledge graph embedding via
  dynamic mapping matrix,'' in {\em Proceedings of the 53rd annual meeting of
  the association for computational linguistics and the 7th international joint
  conference on natural language processing (volume 1: Long papers)},
  pp.~687--696, 2015.

\bibitem{chao2020pairre}
L.~Chao, J.~He, T.~Wang, and W.~Chu, ``Pairre: Knowledge graph embeddings via
  paired relation vectors,'' {\em arXiv preprint arXiv:2011.03798}, 2020.

\bibitem{balazevic2019multi}
I.~Balazevic, C.~Allen, and T.~Hospedales, ``Multi-relational poincar{\'e}
  graph embeddings,'' {\em Advances in Neural Information Processing Systems},
  vol.~32, 2019.

\bibitem{ebisu2019generalized}
T.~Ebisu and R.~Ichise, ``Generalized translation-based embedding of knowledge
  graph,'' {\em IEEE Transactions on Knowledge and Data Engineering}, vol.~32,
  no.~5, pp.~941--951, 2019.

\bibitem{nickel2011three}
M.~Nickel, V.~Tresp, H.-P. Kriegel, {\em et~al.}, ``A three-way model for
  collective learning on multi-relational data.,'' in {\em Icml}, vol.~11,
  pp.~3104482--3104584, 2011.

\bibitem{yang2014embedding}
B.~Yang, W.-t. Yih, X.~He, J.~Gao, and L.~Deng, ``Embedding entities and
  relations for learning and inference in knowledge bases,'' {\em arXiv
  preprint arXiv:1412.6575}, 2014.

\bibitem{liu2017analogical}
H.~Liu, Y.~Wu, and Y.~Yang, ``Analogical inference for multi-relational
  embeddings,'' in {\em International conference on machine learning},
  pp.~2168--2178, PMLR, 2017.

\bibitem{trouillon2016complex}
T.~Trouillon, J.~Welbl, S.~Riedel, {\'E}.~Gaussier, and G.~Bouchard, ``Complex
  embeddings for simple link prediction,'' in {\em International conference on
  machine learning}, pp.~2071--2080, PMLR, 2016.

\bibitem{nickel2016holographic}
M.~Nickel, L.~Rosasco, and T.~Poggio, ``Holographic embeddings of knowledge
  graphs,'' in {\em Proceedings of the AAAI conference on artificial
  intelligence}, vol.~30, 2016.

\bibitem{cao2021dual}
Z.~Cao, Q.~Xu, Z.~Yang, X.~Cao, and Q.~Huang, ``Dual quaternion knowledge graph
  embeddings,'' in {\em Proceedings of the AAAI conference on artificial
  intelligence}, vol.~35, pp.~6894--6902, 2021.

\bibitem{guo2021bique}
J.~Guo and S.~Kok, ``Bique: Biquaternionic embeddings of knowledge graphs,''
  {\em arXiv preprint arXiv:2109.14401}, 2021.

\bibitem{liang2024effective}
Q.~Liang, W.~Wang, J.~Yu, and F.~Bao, ``Effective knowledge graph embedding
  with quaternion convolutional networks,'' in {\em CCF International
  Conference on Natural Language Processing and Chinese Computing},
  pp.~183--196, Springer, 2024.

\bibitem{gao2021quatde}
H.~Gao, K.~Yang, Y.~Yang, R.~Y. Zakari, J.~W. Owusu, and K.~Qin, ``Quatde:
  Dynamic quaternion embedding for knowledge graph completion,'' {\em arXiv
  preprint arXiv:2105.09002}, 2021.

\bibitem{liang2024hierarchy}
Q.~Liang, W.~Wang, J.~Yu, and F.~Bao, ``Hierarchy-aware quaternion embedding
  for knowledge graph completion,'' in {\em 2024 International Joint Conference
  on Neural Networks (IJCNN)}, pp.~1--8, IEEE, 2024.

\bibitem{li2022quatse}
J.~Li, X.~Su, X.~Ma, and G.~Gao, ``Quatse: Spherical linear interpolation of
  quaternion for knowledge graph embeddings,'' in {\em CCF International
  Conference on Natural Language Processing and Chinese Computing},
  pp.~209--220, Springer, 2022.

\bibitem{bordes2014semantic}
A.~Bordes, X.~Glorot, J.~Weston, and Y.~Bengio, ``A semantic matching energy
  function for learning with multi-relational data: Application to word-sense
  disambiguation,'' {\em Machine learning}, vol.~94, pp.~233--259, 2014.

\bibitem{dong2014knowledge}
X.~Dong, E.~Gabrilovich, G.~Heitz, W.~Horn, N.~Lao, K.~Murphy, T.~Strohmann,
  S.~Sun, and W.~Zhang, ``Knowledge vault: A web-scale approach to
  probabilistic knowledge fusion,'' in {\em Proceedings of the 20th ACM SIGKDD
  international conference on Knowledge discovery and data mining},
  pp.~601--610, 2014.

\bibitem{dettmers2018convolutional}
T.~Dettmers, P.~Minervini, P.~Stenetorp, and S.~Riedel, ``Convolutional 2d
  knowledge graph embeddings,'' in {\em Proceedings of the AAAI conference on
  artificial intelligence}, vol.~32, 2018.

\bibitem{nguyen2017novel}
D.~Q. Nguyen, T.~D. Nguyen, D.~Q. Nguyen, and D.~Phung, ``A novel embedding
  model for knowledge base completion based on convolutional neural network,''
  {\em arXiv preprint arXiv:1712.02121}, 2017.

\bibitem{schlichtkrull2018modeling}
M.~Schlichtkrull, T.~N. Kipf, P.~Bloem, R.~Van Den~Berg, I.~Titov, and
  M.~Welling, ``Modeling relational data with graph convolutional networks,''
  in {\em The semantic web: 15th international conference, ESWC 2018,
  Heraklion, Crete, Greece, June 3--7, 2018, proceedings 15}, pp.~593--607,
  Springer, 2018.

\bibitem{vashishth2019composition}
S.~Vashishth, S.~Sanyal, V.~Nitin, and P.~Talukdar, ``Composition-based
  multi-relational graph convolutional networks,'' {\em arXiv preprint
  arXiv:1911.03082}, 2019.

\bibitem{yao2019kg}
L.~Yao, C.~Mao, and Y.~Luo, ``Kg-bert: Bert for knowledge graph completion,''
  {\em arXiv preprint arXiv:1909.03193}, 2019.

\bibitem{hamilton1844ii}
W.~R. Hamilton, ``Ii. on quaternions; or on a new system of imaginaries in
  algebra,'' {\em The London, Edinburgh, and Dublin Philosophical Magazine and
  Journal of Science}, vol.~25, no.~163, pp.~10--13, 1844.

\bibitem{kenwright2023survey}
B.~Kenwright, ``A survey on dual-quaternions,'' {\em arXiv preprint
  arXiv:2303.14765}, 2023.

\bibitem{parcollet2018quaternion}
T.~Parcollet, Y.~Zhang, M.~Morchid, C.~Trabelsi, G.~Linar{\`e}s, R.~De~Mori,
  and Y.~Bengio, ``Quaternion convolutional neural networks for end-to-end
  automatic speech recognition,'' {\em arXiv preprint arXiv:1806.07789}, 2018.

\bibitem{glorot2010understanding}
X.~Glorot and Y.~Bengio, ``Understanding the difficulty of training deep
  feedforward neural networks,'' in {\em Proceedings of the thirteenth
  international conference on artificial intelligence and statistics},
  pp.~249--256, JMLR Workshop and Conference Proceedings, 2010.

\bibitem{he2015delving}
K.~He, X.~Zhang, S.~Ren, and J.~Sun, ``Delving deep into rectifiers: Surpassing
  human-level performance on imagenet classification,'' in {\em Proceedings of
  the IEEE international conference on computer vision}, pp.~1026--1034, 2015.

\bibitem{toutanova2015observed}
K.~Toutanova and D.~Chen, ``Observed versus latent features for knowledge base
  and text inference,'' in {\em Proceedings of the 3rd workshop on continuous
  vector space models and their compositionality}, pp.~57--66, 2015.

\bibitem{wang2018knowledge}
Z.~Wang, J.~Zhang, J.~Feng, and Z.~Chen, ``Knowledge embedding for knowledge
  graph completion,'' {\em IEEE Transactions on Knowledge and Data
  Engineering}, vol.~30, no.~12, pp.~2281--2293, 2018.

\bibitem{duchi2011adaptive}
J.~Duchi, E.~Hazan, and Y.~Singer, ``Adaptive subgradient methods for online
  learning and stochastic optimization.,'' {\em Journal of machine learning
  research}, vol.~12, no.~7, 2011.

\end{thebibliography}
\vskip -2\baselineskip plus -1fil
\begin{IEEEbiography}[{\includegraphics[width=1in,height=1.25in,trim={0cm 0cm 0cm 0cm}]{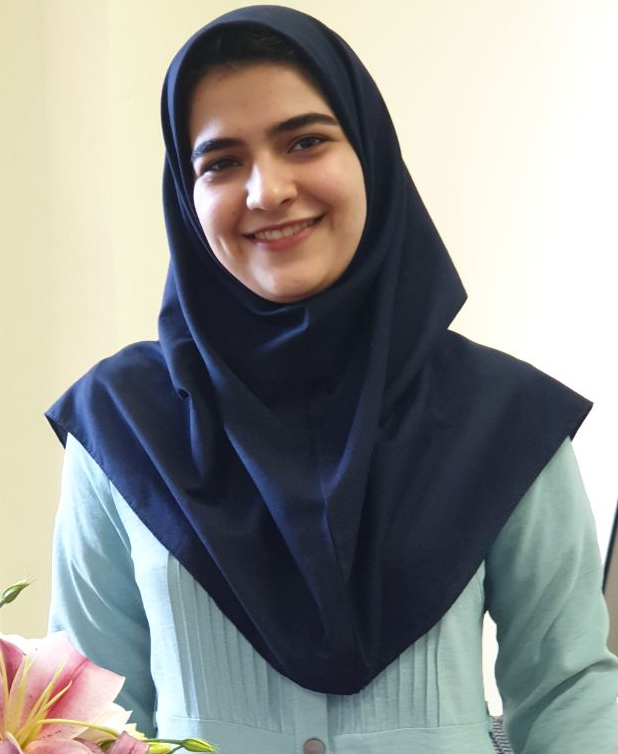}}]{\textbf{Hamideh-Sadat Fazael-Ardakani}}
received her B.Sc. degree in electronic engineering from Yazd university in 2017, Yazd, Iran and M.Sc. degree in communications engineering from Iran University of Science \& Technology, Tehran, Iran in 2020. She is currently pursuing her Ph.D. in the University of Tehran, Tehran, Iran. Her main research interests are Quaternion Signal and Image processing, Quaternion Graph Signal Processing, Quaternion Graph Embedding, compressed sensing and statistical signal processing.
\end{IEEEbiography}
\vskip -2\baselineskip plus -1fil
\begin{IEEEbiography}
	[{\includegraphics[width=1in,height=1.5in,clip,keepaspectratio]{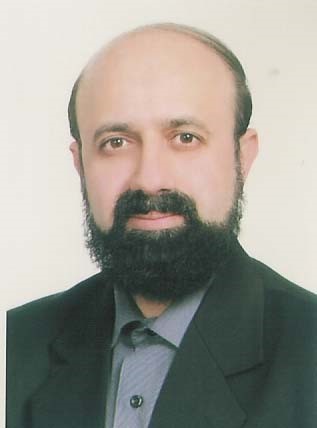}}]{\textbf{Hamid Soltanian-Zadeh, PhD}}
    received BS and MS degrees in electrical engineering: electronics (with honors) from the University of Tehran, Tehran, Iran in 1986 and MSE and PhD degrees in electrical engineering: systems and bioelectrical sciences from the University of Michigan, Ann Arbor, Michigan, USA, in 1990 and 1992, respectively. Since 1988, he has been with the Department of Radiology, Henry Ford Health System, Detroit, Michigan, USA, where he is currently a Senior Scientist and the director of the Medical Image Analysis Lab. Since 1994, he has also been with the School of Electrical and Computer Engineering, the University of Tehran, Tehran, Iran, where he is currently a full Professor and the director of the Biomedical Engineering Lab. Prof. Soltanian-Zadeh is also affiliated with and has active research collaborations with the Wayne State University, Detroit, MI, USA and the Institute for Research in Fundamental Sciences (IPM), Tehran, Iran. His research interests include medical imaging, signal and image processing and analysis, pattern recognition, and neural networks.   He has co-authored over 1000 papers in journals and conference records or as book chapters in these areas. Several of his presentations received honorable mention awards at the SPIE and IEEE conferences. He has delivered numerous invited lectures nationally and internationally. In addition to the SPIE where he is a fellow, he is a senior member of IEEE and a member of ISMRM, ISBME, and ISMVIP and has served on the scientific committees of several national and international conferences and review boards of over 60 scientific journals. Dr. Soltanian-Zadeh is an associate member of the Iranian Academy of Sciences and has served on the study sections of the National Institutes of Health (NIH), National Science Foundation (NSF), American Institute of Biological Sciences (AIBS), and international funding agencies.
\end{IEEEbiography}
%

\vspace{11pt}



\end{document}